\PassOptionsToPackage{hyphens}{url}
\documentclass{bmvc2k}

\title{HSMLA: Hierarchical Softmax Multi-scale Linear Attention for Efficient Vision Transformers}

\addauthor{Dong Liu}{pikeliu@ucla.edu}{1,*}
\addauthor{Yanxuan Yu}{yy3523@columbia.edu}{2,*}
\addauthor{Renata Borovica-Gajic}{renata.borovica@unimelb.edu.au}{3}
\addauthor{Tong Geng}{tony.geng@rice.edu}{4}
\addauthor{Ying Nian Wu}{ywu@stat.ucla.edu}{1,\dag}

\addinstitution{University of California, Los Angeles, USA}
\addinstitution{Columbia University, USA}
\addinstitution{The University of Melbourne, Australia}
\addinstitution{Rice University, USA}

\runninghead{Liu et al.}{HSMLA}

\usepackage{amssymb}
\usepackage{amsfonts}
\usepackage{algorithm}
\usepackage{algorithmic}
\usepackage{booktabs}
\usepackage{float}
\usepackage{multirow,multicol}

\begin{document}

\maketitle
{\renewcommand{\thefootnote}{}\footnotetext{\mbox{$^{*}$ Equal contribution.\ $^{\dagger}$ Corresponding author.}}}

\begin{abstract}
Vision transformers face significant computational overheads in high-resolution dense prediction due to the quadratic complexity of self-attention. Linear attention offers efficiency but sacrifices local context modeling. We propose \textbf{HSMLA (Hierarchical Softmax Multi-scale Linear Attention)}, which combines ReLU-based linear attention for global context, selective softmax refinement for critical local features, and multi-scale token representations via depthwise convolutions. HSMLA achieves superior accuracy–efficiency trade-offs: up to $4.2\times$ inference-time speedup across dense prediction tasks, $87.3\%$ Dice with $3.2\times$ speedup on CT organ segmentation, and $94.2\%$ AUC with $4.1\times$ speedup on pathology WSI.

\end{abstract}

\section{Introduction}

Vision Transformers (ViTs) have achieved remarkable results across a wide range of vision tasks~\cite{dosovitskiy2020image,deit,swin}, but the quadratic complexity of self-attention limits their applicability to high-resolution dense prediction~\cite{swin,segformer,segnext}. Linear attention~\cite{katharopoulos2020transformers,choromanski2020performer} offers $\mathcal{O}(N)$ complexity but produces flatter attention distributions, weakening local context modeling (Figure~\ref{fig:arch}). This is particularly critical in medical imaging, where organ boundaries and tumor margins require sharp, localized attention.

We propose \textbf{HSMLA (Hierarchical Softmax Multi-scale Linear Attention)} (see Figure~\ref{fig:arch}), which combines multi-scale ReLU-based linear attention for efficient global context aggregation with query-dependent selective softmax attention on a sparse set of critical tokens for local refinement. Multi-scale token representations via depthwise convolutions and an FFN--DWConv sandwich structure further enhance representational diversity and local modeling. By routing only a small fraction of queries to full softmax attention while the remaining queries use linear attention, HSMLA achieves approximately $3\times$ inference-time speedup over standard self-attention while maintaining accuracy.

Compared to EfficientViT~\cite{cai2023efficientvit}, which employs multi-scale linear attention but applies it uniformly across tokens, HSMLA introduces content-aware selective softmax refinement that recovers sharp local structures important for dense prediction. Unlike dynamic or sparse ViTs that focus on token pruning for classification~\cite{rao2021dynamicvit,yin2022vit,chen2023sparsevit}, HSMLA is designed for high-resolution dense prediction and medical imaging, combining query-sparse softmax with multi-scale linear attention.

High-resolution vision tasks make these constraints particularly severe. Semantic segmentation in autonomous driving typically operates on megapixel-scale street scenes; single-image super-resolution often uses large 4K inputs; and biomedical applications such as CT, MRI, and whole-slide imaging (WSI) involve volumetric or gigapixel data. In many of these settings, especially for portable and intraoperative systems, models must run under tight latency and memory budgets on edge devices (e.g., Jetson-class GPUs) while preserving fine structural details. This motivates an attention mechanism that allocates full softmax computation only where it is most beneficial, and relies on efficient linear attention elsewhere.

Experiments on Cityscapes~\cite{cordts2016cityscapes}, DIV2K~\cite{timofte2017ntire}, and multiple biomedical imaging benchmarks show that HSMLA achieves up to $4\times$ inference-time speedup with state-of-the-art accuracy. Our main contributions are:
(i) an analysis of the limitations of linear attention for dense prediction;
(ii) HSMLA, a hierarchical attention module that combines multi-scale linear attention with query-sparse selective softmax gating;
(iii) extensive experiments demonstrating superior accuracy--efficiency trade-offs on semantic segmentation, super-resolution, and clinical imaging tasks.

\section{Related Work}
Transformers for vision have rapidly evolved along several complementary directions, including efficient architectures for high-resolution inputs, linear and sparse attention mechanisms, and multi-scale or dynamic routing strategies. In this section, we first review efficient Vision Transformers that target reduced computational and memory cost, then discuss linear attention variants and their limitations for dense prediction, and finally summarize multi-scale and dynamic attention methods that adapt computation across tokens and resolutions.

\subsection{Efficient Vision Transformers}
ViTs~\cite{dosovitskiy2020image} achieve state-of-the-art performance with long-range modeling~\cite{deit,swin,liu2022convnext}, but quadratic complexity limits high-resolution dense prediction~\cite{wang2021pyramid}. Swin~\cite{swin} uses shifted window attention ($\mathcal{O}(W^2 N)$); SegFormer~\cite{segformer} adopts hierarchical MLP decoders; PVT~\cite{wang2021pyramid} employs spatial-reduction; MobileViT~\cite{mehta2021mobilevit} integrates local convolutions; EfficientViT~\cite{cai2023efficientvit}, FasterViT~\cite{hatamizadeh2024fastervit}, and VMamba~\cite{liu2024vmamba} explore hardware-efficient designs. Most still face bottlenecks from softmax and tensor reshaping on edge hardware.

\subsection{Linear Attention Mechanisms}
Linear attention offers $\mathcal{O}(N)$ alternatives: Performers~\cite{choromanski2020performer} use random Fourier features; Linformer~\cite{wang2020linformer} projects keys/values; Reformer~\cite{kitaev2020reformer} uses locality-sensitive hashing; works~\cite{katharopoulos2020transformers,shen2021efficient} adopt ReLU/ELU kernels. EfficientViT~\cite{cai2023efficientvit} combines ReLU linear attention with multi-scale tokens. Complementary efficiency techniques reuse or selectively retain activations via caching~\cite{liu2025fastcache,liu2026adacorrection,liu2026accelerating,liu2025tinyserve,liu2025pikv,liu2026kvlearn} and hardware-aware attention~\cite{wang2020hat,wang2021spatten,liu2026mka}. However, linear attention produces flatter distributions~\cite{xiong2021nystromformer,peng2021random}, limiting sharp local details for dense prediction~\cite{milakov2018online}. Sparse methods~\cite{zaheer2020bigbird,beltagy2020longformer,child2019generating} use structured sparsity but lack content-aware gating.

\subsection{Multi-Scale and Dynamic Attention}
Multi-scale mechanisms integrate context across resolutions~\cite{zhao2017pyramid,chen2017deeplab}. PVT~\cite{wang2021pyramid}, SegNeXt~\cite{segnext}, PSA~\cite{zhao2018psanet}, and BiFormer~\cite{zhu2023biformer} employ hierarchical or dynamic routing, while DynamicViT~\cite{rao2021dynamicvit}, EViT~\cite{liang2022not}, SparseViT~\cite{chen2023sparsevit}, AdaViT~\cite{meng2022adavit}, A-ViT~\cite{yin2022vit}, and Evo-ViT~\cite{xu2022evo} adapt computation via pruning, depth/width scaling, early exiting, or token merging. In medical imaging, TransUNet~\cite{chen2021transunet} and SwinUNet~\cite{cao2021swinunet} bring transformers into U-Net-style CT/MRI segmentation, and efficient biomedical designs~\cite{yu2026coprimeeeg,guo2024reconformer,valanarasu2021medical} further reduce sensing and compute cost; yet medical ViTs typically retain dense softmax attention, and efficient backbones such as EfficientViT~\cite{cai2023efficientvit}, FasterViT~\cite{hatamizadeh2024fastervit} and EfficientFormer~\cite{li2022efficientformer} still lack query-sparse local refinement for precise organ boundaries and tumor margins under tight budgets. Our work bridges this gap by combining multi-scale linear attention with content-aware, query-sparse softmax refinement for efficient yet locally precise high-resolution medical modeling.

\begin{figure}[h]
\centering
\subfigure[Attention patterns: Softmax (sharp) vs.\ Linear (flat) vs.\ HSMLA (hierarchical)]{%
    \label{fig:arch}\includegraphics[width=0.48\linewidth]{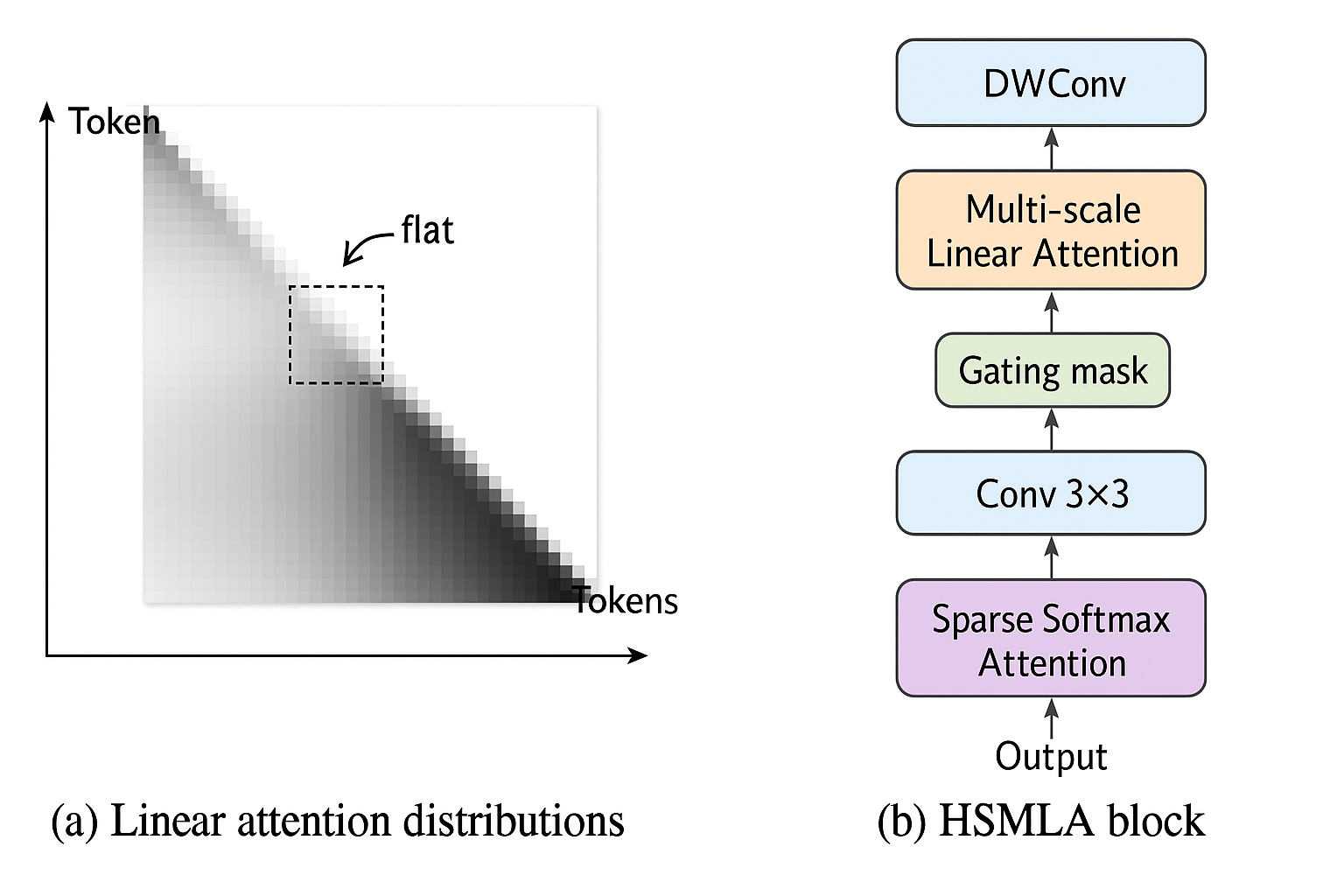}}
\hfill
\subfigure[HSMLA block architecture]{%
    \label{fig:model}\includegraphics[width=0.48\linewidth]{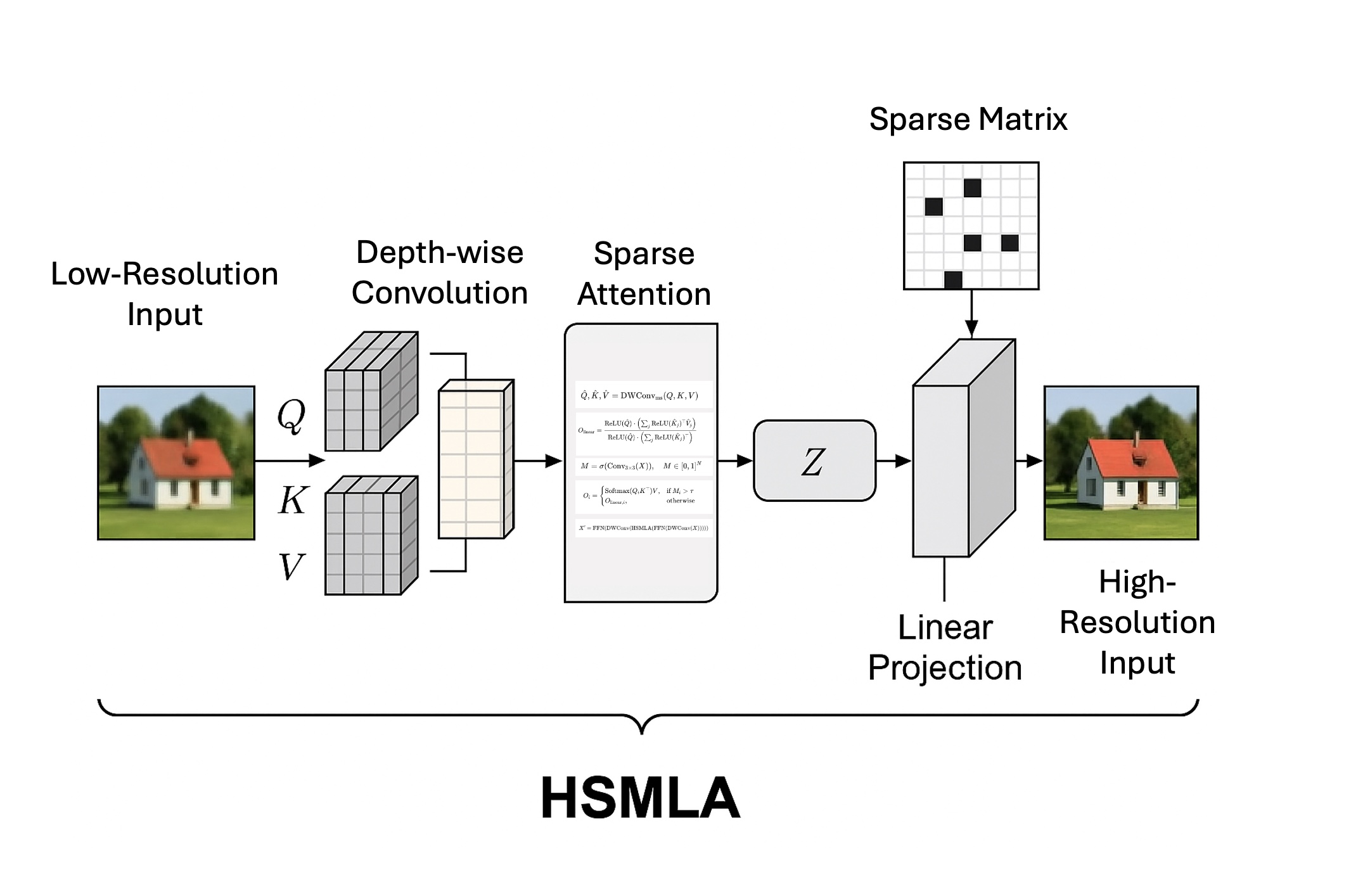}}
\caption{HSMLA architecture: (a) attention pattern comparison; (b) block design.}
\end{figure}

We propose HSMLA to combine linear attention with selective softmax at critical regions, retaining global efficiency while recovering local expressiveness (Figure~\ref{fig:arch}).

\section{Methodology}
\label{sec:method}

We present \textbf{HSMLA} (Figure~\ref{fig:model}), which combines ReLU-based linear attention for efficient global context aggregation with sparse softmax refinement for locally sharp modeling. HSMLA is built on the principle of allocating expensive softmax attention only to a sparse set of critical \emph{spatial blocks}, while using efficient linear attention everywhere else. In this section, we first describe the multi-scale ReLU-based linear attention backbone that provides global context at $\mathcal{O}(N)$ complexity. We then introduce a hierarchical, content-aware gating mechanism that identifies critical blocks and selectively applies local softmax attention for refinement. Next, we formulate the two-path residual kernel that decouples global linear aggregation from sparse local correction, describe a hardware-aware design for tile-compacted inference and gate regularisation, and detail how HSMLA is integrated into a sandwich-style transformer block. We conclude with a complexity analysis and pseudocode for a complete HSMLA layer.

\subsection{Multi-Scale ReLU Linear Attention}

Inspired by EfficientViT~\cite{cai2023efficientvit}, we adopt ReLU-based linear attention to efficiently compute global context. Given an input sequence $X \in \mathbb{R}^{N \times d}$, where $N$ is the number of tokens and $d$ is the embedding dimension, we first project $X$ to queries, keys, and values:
\begin{equation}
Q, K, V \in \mathbb{R}^{N \times d} = \text{LinearProjection}(X).
\end{equation}
For dense prediction, tokens correspond to positions on a 2D grid; we therefore reshape $Q, K, V$ to $H \times W \times d$ (with $N = HW$) and apply multi-scale depthwise convolutions (DWConv$_\text{ms}$) to enrich local context at multiple receptive fields:
\begin{align}
\hat{Q} &= \text{DWConv}_\text{ms}(Q), \\
\hat{K} &= \text{DWConv}_\text{ms}(K), \\
\hat{V} &= \text{DWConv}_\text{ms}(V),
\end{align}
where DWConv$_\text{ms}$ uses kernel sizes $\{3 \times 3, 5 \times 5, 7 \times 7\}$ fused by element-wise addition:
\begin{align}
\hat{Q}_i &= \sum_{k \in \{3,5,7\}} \text{DWConv}_{k \times k}(Q)_i, \\
\hat{K}_i &= \sum_{k \in \{3,5,7\}} \text{DWConv}_{k \times k}(K)_i, \\
\hat{V}_i &= \sum_{k \in \{3,5,7\}} \text{DWConv}_{k \times k}(V)_i,
\end{align}
with $\hat{Q}_i, \hat{K}_i, \hat{V}_i \in \mathbb{R}^{d}$ denoting the features at token $i$.

ReLU-based linear attention exploits associativity to reduce the complexity of self-attention from $\mathcal{O}(N^2 d)$ to $\mathcal{O}(N d^2)$. Defining the feature map $\phi(x) = \text{ReLU}(x)$ and the global aggregation terms
\begin{align}
Z &= \sum_{j=1}^{N} \phi(\hat{K}_j)^\top \hat{V}_j \in \mathbb{R}^{d \times d}, \\
D &= \sum_{j=1}^{N} \phi(\hat{K}_j)^\top \in \mathbb{R}^{d},
\end{align}
the linear attention output for token $i$ is
\begin{equation}
O^{\text{linear}}_i = \frac{\phi(\hat{Q}_i) Z}{\phi(\hat{Q}_i) D} \in \mathbb{R}^{d}.
\end{equation}
This formulation computes $Z$ and $D$ once per layer and reuses them for all queries, yielding linear-time dependence on $N$.

\subsection{Hierarchical Block-Level Gating}
\label{sec:gate}

While linear attention captures global context efficiently, it tends to produce flatter attention distributions and lacks local sharpness (Figure~\ref{fig:arch}). To selectively recover sharp local structures, we introduce a learned gating mask $M \in [0,1]^T$ over $T$ spatial blocks that determines which regions receive full softmax attention. Intuitively, blocks with higher gate values are more likely to correspond to boundaries, fine textures, or other high-complexity regions and therefore benefit from sharp local refinement.

Rather than scoring each token independently, we adopt a \emph{hierarchical} gating mechanism that operates at two levels: the \emph{head level}, where each attention head assesses criticality from its own feature perspective, and the \emph{block level}, where routing decisions are made over contiguous spatial tiles rather than individual tokens.

\paragraph{Head-level scores.}
For each head $h \in \{1,\dots,H\}$ and each spatial block $b$, we compute a head-specific gating score from the multi-scale features:
\begin{equation}
\label{eq:head_gate}
G^{(h)}_b = \sigma\!\left(\mathbf{w}^{(h)\top} \text{AvgPool}\!\left(\text{DWConv}_{3\times3}\!\left([\hat{Q}^{(h)}_b;\, \hat{K}^{(h)}_b;\, \hat{V}^{(h)}_b]\right)\right) + b^{(h)}\right),
\end{equation}
where $[\cdot;\cdot;\cdot]$ denotes concatenation along the channel dimension, and $\mathbf{w}^{(h)}, b^{(h)}$ are learnable parameters. This allows each head to focus on different aspects of local structure (e.g., edges, textures, or semantic boundaries).

\paragraph{Block-level aggregation.}
We aggregate the head-wise scores into a shared block-level gate:
\begin{equation}
\label{eq:block_gate}
M_b = \frac{1}{H}\sum_{h=1}^{H} G^{(h)}_b \;=\; \sigma\!\left(\text{AvgPool}\!\left(\text{Conv}_{3\times 3}(X_b)\right)\right) \in [0,1],
\end{equation}
which is the mask used in Eq.~\eqref{eq:diff_mixture} and in the inference-time routing below. Token-level gates would produce scattered activation patterns that are expensive on modern hardware due to non-contiguous memory access and per-token branch divergence. Operating at the block level means all $B^2$ tokens inside block $b$ share the same routing decision $M_b$, so the selected set
\begin{equation}
\mathcal{B}_\tau = \{b : M_b > \tau\}
\end{equation}
is a contiguous collection of tiles. This design enables: \textbf{(i)} coalesced memory access, since all keys/values for a tile reside in a contiguous region; \textbf{(ii)} no per-token branch divergence, as CUDA warps execute uniformly within each tile; and \textbf{(iii)} Tensor Core compatibility, since each tile's local softmax is a dense $w^2 \times w^2$ matrix multiply.

Let $\alpha$ denote the fraction of blocks that satisfy $M_b > \tau$:
\begin{equation}
\alpha = \frac{|\mathcal{B}_\tau|}{T} = \frac{1}{T}\sum_{b=1}^{T} \mathbf{1}[M_b > \tau].
\end{equation}
In practice, we choose $\tau$ such that $\alpha \approx 0.3$, i.e., only about $30\%$ of blocks invoke local softmax refinement at inference, while the remaining $70\%$ rely solely on linear attention.

\paragraph{Per-head output routing.}
For completeness, the per-head output at inference can be written as
\begin{equation}
O^{(h)}_i =
\begin{cases}
O^{\text{local},(h)}_i + \bigl(O^{\text{linear},(h)}_i - O^{\text{lin\text{-}local},(h)}_i\bigr), & G^{(h)}_{\beta(i)} > \tau, \\[4pt]
O^{\text{linear},(h)}_i, & \text{otherwise},
\end{cases}
\end{equation}
where $d_h = d/H$ and $Z^{(h)}, D^{(h)}$ are the head-specific linear-attention aggregates. In practice, we use the shared block mask $M_b$ to route the full multi-head output via the two-path residual formulation described next.

\subsection{Hierarchical Sparse Softmax Refinement: Two-Path Residual Kernel}
\label{sec:twopath}

Given the block-level gate $M_b$, HSMLA routes selected regions to a local softmax path while keeping efficient linear attention as the global backbone. A naive mixture that replaces linear output with full-sequence softmax,
$O^{\text{final}}_i = M_i \cdot \text{SoftmaxAttn}(Q_i, K, V) + (1 - M_i) \cdot O^{\text{linear}}_i$,
discards global context for gated tokens and incurs $\mathcal{O}(\alpha N^2 d)$ cost. We instead restrict softmax to a local neighbourhood and formulate refinement as a sparse residual correction on top of the global linear path.

\paragraph{Local neighbourhood.}
For each token $i$ in a selected block, let $\mathcal{N}(i)$ denote its local window of size $w \times w$ (e.g., $7\!\times\!7$ or $11\!\times\!11$). The local softmax output is
\begin{equation}
\label{eq:local_softmax}
O^{\text{local}}_i = \text{SoftmaxAttn}\!\left(Q_i,\, K_{\mathcal{N}(i)},\, V_{\mathcal{N}(i)}\right),
\end{equation}
where $K_{\mathcal{N}(i)}, V_{\mathcal{N}(i)} \in \mathbb{R}^{w^2 \times d}$ are the keys and values restricted to the $w^2$ neighbours. This replaces the original formulation that attended over all $N$ tokens, reducing the softmax step from $\mathcal{O}(\alpha N^2 d)$ to $\mathcal{O}(\alpha N w^2 d)$.

We also compute the linear attention output restricted to the same local window:
\begin{equation}
O^{\text{lin\text{-}local}}_i = \frac{\phi(\hat{Q}_i)\, Z_{\mathcal{N}(i)}}{\phi(\hat{Q}_i)\, D_{\mathcal{N}(i)}},
\end{equation}
where $Z_{\mathcal{N}(i)} = \sum_{j \in \mathcal{N}(i)} \phi(\hat{K}_j)^\top \hat{V}_j$ and $D_{\mathcal{N}(i)} = \sum_{j \in \mathcal{N}(i)} \phi(\hat{K}_j)^\top$.

\paragraph{Two-path residual formulation.}
Rather than binary-routing tokens between softmax and linear attention, HSMLA adopts a \emph{residual refinement} decomposition:
\begin{equation}
\label{eq:two_path}
O_i = O^{\text{linear}}_i + R_i,
\end{equation}
where the refinement residual $R_i$ is non-zero only for selected blocks:
\begin{equation}
\label{eq:residual}
R_i = \mathbf{1}[M_{\beta(i)} > \tau]\;\bigl(O^{\text{local}}_i - O^{\text{lin\text{-}local}}_i\bigr),
\end{equation}
and $\beta(i)$ denotes the block containing token $i$. Intuitively, $O^{\text{linear}}_i$ aggregates \emph{global} context for every token, so the backbone costs $\mathcal{O}(N d^2)$; $R_i$ then adds only the \emph{local correction} that softmax provides beyond linear attention, incurring an additional $\mathcal{O}(\alpha N w^2 d)$ over the $\alpha$-fraction of gated blocks, and this correction is a sparse residual that can be scheduled as a separate kernel.

To better exploit the complementary strengths of both attention mechanisms, we formulate the refinement as a residual correction rather than a binary replacement: every token retains its full global context from the linear path, while the sparse softmax pass contributes only the local correction $R_i$ where it matters most. This also narrows the training–inference gap, since the differentiable soft version $R_i = M_{\beta(i)}\,(O^{\text{local}}_i - O^{\text{lin\text{-}local}}_i)$ converges smoothly to the hard routing at inference. Furthermore, because $R_i = 0$ for non-selected tokens, the refinement pass is a sparse residual that can be launched as a separate, densely-packed CUDA kernel over only $|\mathcal{B}_\tau| \cdot B^2$ tokens, decoupled from the linear attention path.

\paragraph{Training-time differentiable mixture.}
During training, we use a differentiable mixture of softmax and linear attention for stability, replacing the hard indicator with the soft gate $M_{\beta(i)} \in [0,1]$:
\begin{equation}
\label{eq:diff_mixture}
O^{\text{final}}_i = O^{\text{linear}}_i + M_{\beta(i)}\;\bigl(O^{\text{local}}_i - O^{\text{lin\text{-}local}}_i\bigr),
\end{equation}
where $\text{SoftmaxAttn}(Q_i, K_{\mathcal{N}(i)}, V_{\mathcal{N}(i)})$ in Eq.~\eqref{eq:local_softmax} denotes standard multi-head self-attention restricted to the local neighbourhood. At inference, we apply hard thresholding with threshold $\tau$ and route only blocks in $\mathcal{B}_\tau$ to the softmax path:
\begin{equation}
\label{eq:inference_routing}
O^{\text{final}}_i =
\begin{cases}
O^{\text{linear}}_i + O^{\text{local}}_i - O^{\text{lin\text{-}local}}_i, & M_{\beta(i)} > \tau, \\[4pt]
O^{\text{linear}}_i, & \text{otherwise},
\end{cases}
\end{equation}
which recovers the two-path decomposition in Eq.~\eqref{eq:two_path}. To see this, substitute Eq.~\eqref{eq:residual} into Eq.~\eqref{eq:two_path}: for tokens in selected blocks ($M_{\beta(i)} > \tau$), the indicator $\mathbf{1}[M_{\beta(i)} > \tau] = 1$ and
\begin{equation}
O^{\text{final}}_i = O^{\text{linear}}_i + \bigl(O^{\text{local}}_i - O^{\text{lin\text{-}local}}_i\bigr),
\end{equation}
which is exactly the first branch of Eq.~\eqref{eq:inference_routing}; for all other tokens, $\mathbf{1}[M_{\beta(i)} > \tau] = 0$ so $R_i = 0$ and $O^{\text{final}}_i = O^{\text{linear}}_i$. Thus, hard thresholding at inference simply binarises the soft gate $M_{\beta(i)}$ used during training, while preserving the same residual structure: the global linear path is always computed for every token, and the local softmax correction is added only where the gate activates.

\subsection{Hardware-Aware Design}
\label{sec:hw_design}

The gating mechanism in Eq.~\eqref{eq:block_gate} is learned purely from task supervision, which does not penalise fragmented or over-dense activation patterns that are inefficient on hardware. We therefore co-design the inference pipeline (below) with two training regularisers that encourage hardware-friendly activation patterns:
\begin{equation}
\label{eq:total_loss}
\mathcal{L} = \mathcal{L}_{\text{task}} + \lambda_1 \left|\frac{1}{T}\sum_{b=1}^{T} M_b - \rho\right| + \lambda_2\, \mathcal{C}_{\text{hw}}(M),
\end{equation}
where $\rho$ is a learnable target softmax budget and

\begin{equation}
\label{eq:spatial_reg}
\mathcal{C}_{\text{hw}}(M) = \sum_{\substack{b, b' \,\text{adjacent}}} \!\!\!|M_b - M_{b'}|.
\end{equation}

The first regulariser (budget term) penalises deviations of the mean activation from the learnable target ratio $\rho$, ensuring predictable FLOP and memory traffic at inference. The second regulariser (spatial smoothness term) penalises abrupt changes between neighbouring blocks, encouraging $\mathcal{B}_\tau$ to form contiguous tile clusters rather than scattered activations---a pattern that maps poorly to the GPU memory hierarchy (HBM $\rightarrow$ L2 $\rightarrow$ shared memory). We use $\lambda_1 = 0.01$ and $\lambda_2 = 0.005$ throughout all experiments (ablated in Section~\ref{sec:ablation}).

\paragraph{Hardware implementation.}
At inference, HSMLA is mapped to a \emph{two-stage pipeline} on the memory hierarchy (HBM/DRAM $\rightarrow$ L2 $\rightarrow$ on-chip SRAM). \textbf{Stage~1} executes the global linear path in Eq.~\eqref{eq:two_path} for all $N$ tokens. Because $\hat{Q},\hat{K},\hat{V}$ are stored in row-major $H \!\times\! W$ layout, each warp issues coalesced HBM/DRAM reads; the effective bandwidth is
\begin{equation}
\label{eq:hbm_bw}
B_{\text{eff}}^{\text{linear}} \;\approx\; \min\!\bigl(B_{\text{HBM}},\; \eta_{\text{coalesce}}\, B_{\text{L2}}\bigr),
\end{equation}
where $B_{\text{HBM}}$ and $B_{\text{L2}}$ denote peak HBM/DRAM and L2 bandwidth, and $\eta_{\text{coalesce}} \in (0,1]$ captures how much of each cache line is reused across spatial neighbours ($\eta_{\text{coalesce}} \!\rightarrow\! 1$ when tokens within a tile share routing and fall in the same line). \textbf{Stage~2} refines only blocks in $\mathcal{B}_\tau$. After thresholding $M_b > \tau$, activated tiles are compacted via prefix sum into a dense list $\mathcal{L} = \{b_1,\ldots,b_{|\mathcal{B}_\tau|}\}$, replacing $\mathcal{O}(\alpha N)$ irregular gathers from HBM with $\mathcal{O}(|\mathcal{B}_\tau| \cdot w^2 d)$ batched loads into shared memory (on-chip SRAM). Each $b_k \in \mathcal{L}$ launches one thread block that stages the $w \!\times\! w$ neighbourhood $\mathcal{N}(i)$ in SRAM, evaluates Eq.~\eqref{eq:local_softmax} as a dense $w^2 \!\times\! w^2$ GEMM (Tensor-Core), and writes the residual $R_i$ only at gated positions. End-to-end latency decomposes as
\begin{equation}
\label{eq:hw_latency}
T_{\text{HW}} = \underbrace{T_{\text{linear}}(N,d)}_{\text{HBM-bound}} + \underbrace{S(M)\, T_{\text{launch}}}_{\text{wavefront overhead}} + \underbrace{|\mathcal{B}_\tau|\, T_{\text{tile}}(w,d)}_{\text{L2/SRAM-bound}},
\end{equation}
where $S(M)$ is the number of disconnected segments in $\mathcal{L}$ after compaction---each segment triggers an additional kernel wavefront and extra HBM $\rightarrow$ L2 traffic. The smoothness term $\mathcal{C}_{\text{hw}}$ in Eq.~\eqref{eq:spatial_reg} directly reduces $S(M)$, while the budget term stabilises $|\mathcal{B}_\tau| \approx \rho T$; together they align the learned mask with hardware-friendly access patterns (Figure~\ref{fig:kernel}). Peak activation memory is correspondingly bounded by
\begin{equation}
\label{eq:hw_mem}
M_{\text{peak}} = \underbrace{Nd}_{\text{HBM activations}} + \underbrace{d^2}_{\text{linear aggregates}} + \underbrace{|\mathcal{B}_\tau| \cdot w^2 d}_{\text{SRAM tile buffers}},
\end{equation}
which scales with the sparsity budget $\rho$ rather than full-sequence softmax storage.

\subsection{Block Composition of HSMLA}

We adopt the sandwich architecture from EfficientViT~\cite{cai2023efficientvit}, which interleaves HSMLA with depthwise convolutions and a feed-forward network (FFN):
\begin{equation}
\tilde{X}_i = X_i + \text{HSMLA}(\text{LN}(X_i)), \quad
\bar{X}_i = \tilde{X}_i + \text{DWConv}(\text{LN}(\tilde{X}_i)), \quad
X_{i+1} = \bar{X}_i + \text{FFN}(\text{LN}(\bar{X}_i)),
\end{equation}
where LN denotes layer normalization.

\textbf{Complexity.}
Let the feature map have spatial resolution $H \!\times\! W$, so $N = HW$ tokens arranged on a 2D grid.
Routing operates on non-overlapping tiles of size $B \!\times\! B$ ($T = HW/B^2$ blocks, each with $B^2$ tokens sharing one gate $M_b$), while local softmax uses a fixed window of size $w \!\times\! w$ ($w^2$ neighbours per query).
The overall \emph{arithmetic} complexity is
\begin{equation}
\label{eq:complexity}
\mathcal{C}_{\text{HSMLA}} = \underbrace{\mathcal{O}(N d^2)}_{\text{global linear path}} + \underbrace{\mathcal{O}(\alpha N w^2 d)}_{\text{local softmax on gated blocks}},
\end{equation}
where $\alpha = |\mathcal{B}_\tau|/T$ is the fraction of \emph{blocks} (not tokens) selected at threshold $\tau$.
Because $w^2 \ll N$ in practice (e.g.\ $w \in \{7,11\}$ vs.\ $N \in [10^3,10^4]$), the refinement term is reduced by a factor of $w^2/N$ relative to full-sequence softmax ($\mathcal{O}(\alpha N^2 d)$).
The condition $w \ll \sqrt{N}$ is equivalent to $w^2 \ll HW$: each local window covers only a tiny fraction of the map, so global context is delegated to the linear path and softmax is reserved for within-window sharpening.

\subsection{HSMLA Pseudocode}
For clarity and implementation guidance, Algorithm~\ref{algo:hsmla} summarises the forward pass as a hardware-aligned pipeline: batched global linear attention (Stage~1), block-level gating (Stage~2), tile compaction and batched local-softmax kernels (Stages~3--4, inference), and scatter-add of the sparse residual (Stage~5).
During training, the same structure applies except that local attention runs over all queries with soft gates and the regularisers in Eq.~\eqref{eq:total_loss} are accumulated.

\begin{algorithm}[H]
\caption{HSMLA: Hierarchical Softmax Multi-scale Linear Attention (Two-Path)}
\label{algo:hsmla}
\begin{algorithmic}[1]
\REQUIRE Input $X \in \mathbb{R}^{N \times d}$ ($N = HW$), window $w$, routing block size $B$, threshold $\tau$, training\_mode
\STATE $Q, K, V \leftarrow \text{LinearProjection}(X)$
\STATE $\hat{Q}, \hat{K}, \hat{V} \leftarrow \text{DWConv}_\text{ms}(Q,K,V)$
\STATE \COMMENT{\textbf{Stage 1: global linear path (all tokens, no branching)}}
\STATE $Z \leftarrow \sum_{j=1}^{N} \text{ReLU}(\hat{K}_j)^\top \hat{V}_j$, \quad $D \leftarrow \sum_{j=1}^{N} \text{ReLU}(\hat{K}_j)^\top$
\STATE $O^{\text{linear}} \leftarrow \text{ReLU}(\hat{Q}) \cdot Z \oslash \big(\text{ReLU}(\hat{Q}) \cdot D\big)$ \COMMENT{batch matmul over all $N$ tokens}
\STATE \COMMENT{\textbf{Stage 2: block-level gates ($T = HW/B^2$ tiles)}}
\STATE $M_b \leftarrow \sigma\!\left(\text{AvgPool}(\text{Conv}_{3\times3}(X_b))\right)$ for each block $b \in \{1,\ldots,T\}$
\IF{training\_mode}
    \STATE $R \leftarrow \text{BatchedLocalAttn}(\hat{Q}, \hat{K}, \hat{V}, Q, K, V, w)$ \COMMENT{$\mathcal{O}(N w^2 d)$; soft gate over all queries}
    \STATE $O^{\text{final}} \leftarrow O^{\text{linear}} + M_{\beta(\cdot)} \odot R$
    \STATE $\mathcal{L}_{\text{gate}} \leftarrow \lambda_1|\frac{1}{T}\sum_b M_b - \rho| + \lambda_2 \sum_{b \sim b'} |M_b - M_{b'}|$
\ELSE
    \STATE \COMMENT{\textbf{Stage 3--5: tile-compacted sparse refinement (inference only)}}
    \STATE $\mathcal{B}_\tau \leftarrow \{b : M_b > \tau\}$
    \STATE $\mathcal{L} \leftarrow \text{CompactPrefixSum}(\mathcal{B}_\tau)$ \COMMENT{dense list of $|\mathcal{B}_\tau|$ tile indices}
    \STATE $R \leftarrow \text{BatchedLocalSoftmaxKernel}(\mathcal{L}, Q, K, V, \hat{Q}, \hat{K}, \hat{V}, w, B)$
    \STATE \hspace{1.5em}\COMMENT{one CUDA thread block per $b_k \in \mathcal{L}$; stage $(B\!+\!w\!-\!1)^2$ halo tile in SRAM;}
    \STATE \hspace{1.5em}\COMMENT{evaluate $w^2 \!\times\! w^2$ softmax GEMM per query; $R_i \!\leftarrow\! O^{\text{local}}_i \!-\! O^{\text{lin\text{-}local}}_i$}
    \STATE $O^{\text{final}} \leftarrow O^{\text{linear}} + \text{Scatter}(R, \mathcal{B}_\tau)$ \COMMENT{residual correction only at gated tiles}
\ENDIF
\STATE \textbf{return} $O^{\text{final}}$ (and $\mathcal{L}_{\text{gate}}$ during training)
\end{algorithmic}
\end{algorithm}

\vspace{-1em}
\section{Experiments}
\label{sec:experiments}

We empirically evaluate HSMLA on four representative vision settings: semantic segmentation, image classification, single-image super-resolution, and medical imaging. For each setting, we compare against strong transformer and hybrid baselines and report both accuracy and single-image inference latency. Unless otherwise specified, all latency and speedup numbers refer to \emph{inference-time} performance with batch size 1. Speedup is computed as the ratio between the latency of a designated reference baseline (marked as $1.0\times$ in each table) and the latency of the compared model.
\vspace{-1em}
\subsection{Setup}
We evaluate on semantic segmentation (Cityscapes~\cite{cordts2016cityscapes}, ADE20K~\cite{zhou2017scene}), super-resolution (DIV2K~\cite{timofte2017ntire}), classification (ImageNet-1K~\cite{deng2009imagenet}), and medical imaging benchmarks including BTCV~\cite{landman2015miccai}, CAMELYON16~\cite{bejnordi2017camelyon}, EndoVis 2017~\cite{allan2019endovis}, and LUNA16~\cite{setio2017validation}. We compare HSMLA to EfficientViT~\cite{cai2023efficientvit}, SegFormer~\cite{segformer}, BiFormer~\cite{zhu2023biformer}, FasterViT~\cite{hatamizadeh2024fastervit}, TransUNet~\cite{chen2021transunet}, SwinUNet~\cite{cao2021swinunet}, and other state-of-the-art architectures. For a fair comparison, all models use similar backbone scales (e.g., B0/B1/B2 variants) and are evaluated at the same input resolutions as their respective baselines.

\paragraph{Implementation details.}
We train with AdamW ($\beta_1{=}0.9$, $\beta_2{=}0.999$). For ImageNet-1K, we use $224\times224$ crops for 300 epochs, cosine learning-rate decay, and linear warm-up for 20 epochs, with base learning rate $5\times10^{-4}$ scaled linearly with global batch size relative to batch size 512, weight decay $5\times10^{-2}$, and RandAugment, Mixup, and CutMix as in~\cite{deit,liu2022convnext}. For Cityscapes and ADE20K, we follow the SegFormer~\cite{segformer} setup (MMSegmentation-style): random scaling and cropping up to $1024\times1024$, color jitter, and horizontal flipping; we train for $160{,}000$ iterations with polynomial learning-rate decay (power $1.0$), linear warm-up for $1{,}500$ iterations, base learning rate $6\times10^{-5}$, weight decay $1\times10^{-2}$, and an effective batch size of $8$. For DIV2K $4\times$ super-resolution, we minimize the L1 loss with AdamW, cosine annealing from $3\times10^{-4}$ to zero over $300{,}000$ iterations, and progressive patch-based training from $128\times128$ to $384\times384$ high-resolution crops with batch sizes decreased from $64$ to $8$ across stages, following~\cite{zamir2022restormer,chen2024recursive}. Medical models use backbones initialized from ImageNet-1K pretraining where applicable, and are fine-tuned with AdamW, cosine decay, linear warm-up for 10 epochs, learning rate $1\times10^{-4}$, weight decay $1\times10^{-2}$, and 200 training epochs; preprocessing, resampling, and official train/val splits follow the challenge protocols~\cite{landman2015miccai,bejnordi2017camelyon,allan2019endovis,setio2017validation}. Latency is the mean per-image time over 500 runs after 50 warm-up runs, on one NVIDIA GPU (A100 for BTCV, CAMELYON16, and LUNA16; Jetson AGX Orin for EndoVis), batch size 1, FP16 mixed precision, with CUDA synchronization before each timed run.

\subsection{Semantic Segmentation}

\begin{figure}[h]
\centering
\includegraphics[width=0.45\linewidth]{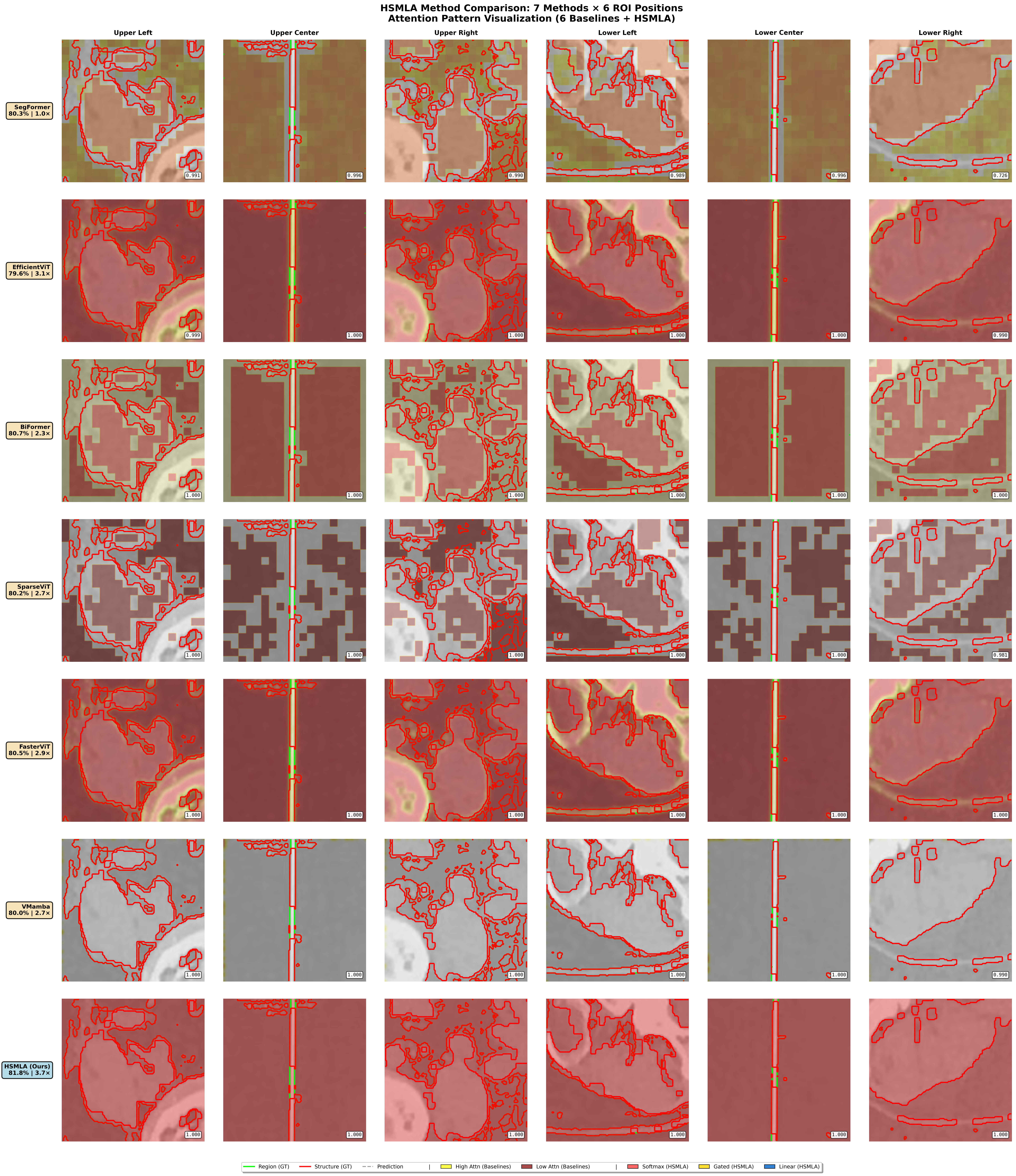}
\caption{Attention pattern comparison across 7 methods and 6 ROI positions. HSMLA (bottom row) demonstrates hierarchical sparse attention: \textcolor{red}{red} regions receive softmax attention for high-complexity features, \textcolor{orange}{yellow} regions use gated linear attention, and \textcolor{blue}{blue} regions employ efficient ReLU linear attention. Baseline methods (top 6 rows) show more uniform or binary attention patterns. HSMLA achieves $81.8\%$ mIoU with $3.7\times$ speedup on Cityscapes, outperforming all baselines.}
\label{fig:method_comparison}
\end{figure}

\begin{figure}[h]
\centering
\subfigure[Performance comparison on semantic segmentation and super-resolution tasks]{%
    \label{fig:performance_comparison}\includegraphics[width=0.45\linewidth]{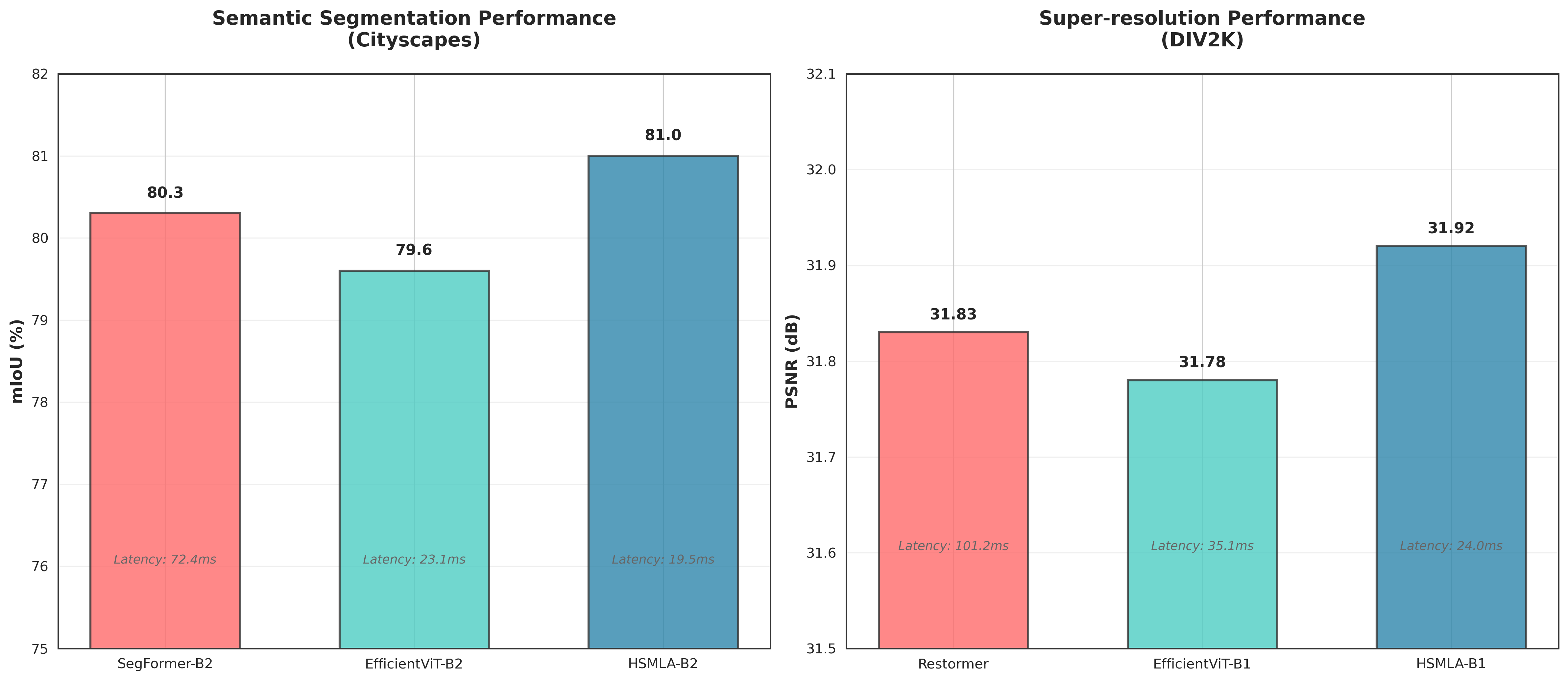}}
\hfill
\subfigure[Runtime breakdown of HSMLA vs.\ baselines]{%
    \label{fig:kernel}\includegraphics[width=0.45\linewidth]{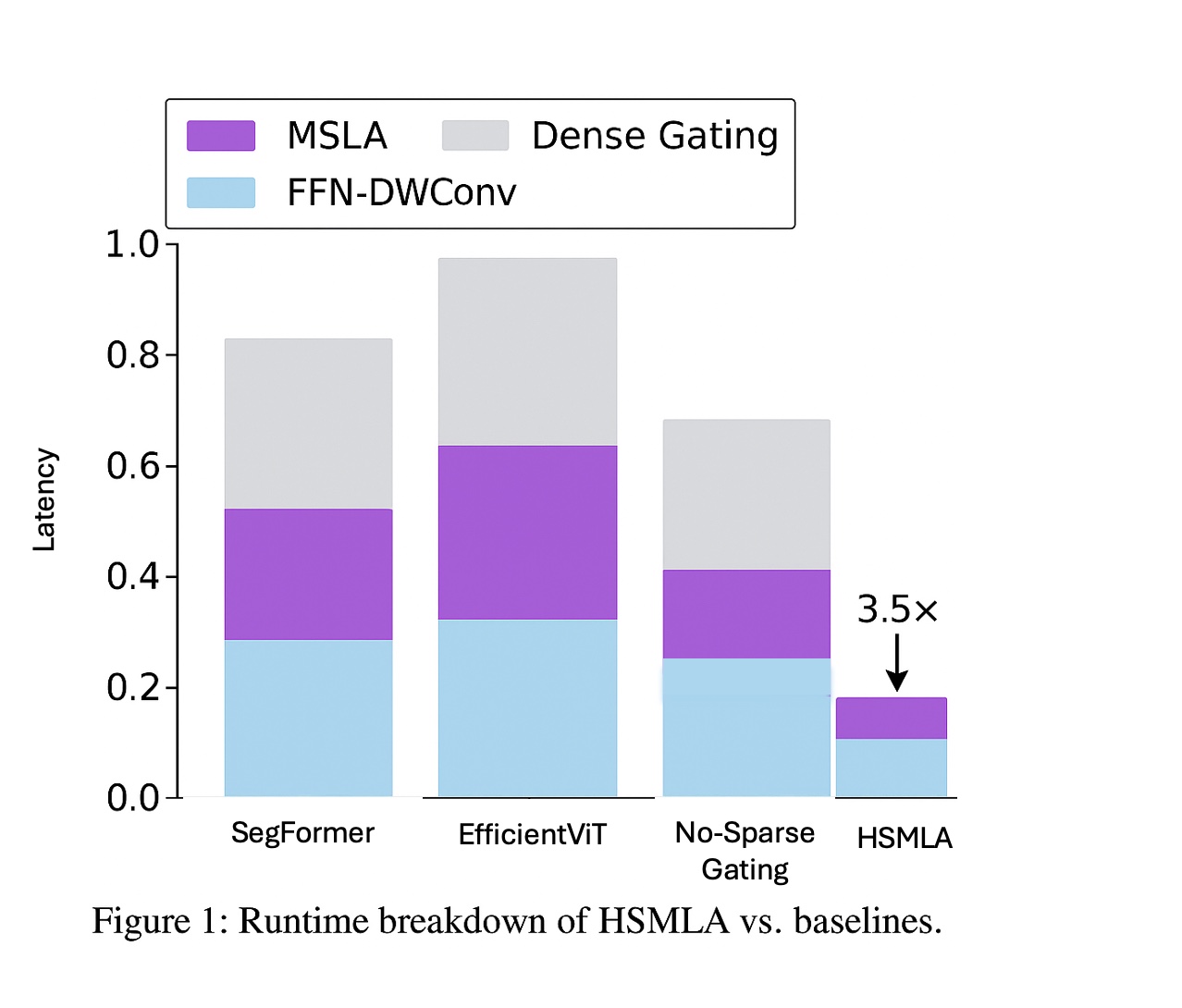}}
\caption{Accuracy and runtime comparison of HSMLA against baseline models. HSMLA achieves a better accuracy–efficiency trade-off by combining multi-scale linear attention with selective softmax refinement.}
\label{fig:performance_and_runtime}
\end{figure}

Figure~\ref{fig:method_comparison} visualizes HSMLA’s hierarchical attention patterns (\textcolor{red}{red}: softmax, \textcolor{orange}{yellow}: gated, \textcolor{blue}{blue}: linear). HSMLA-B2 achieves $81.8\%$ mIoU on Cityscapes with a $3.7\times$ speedup over SegFormer-B2. On ADE20K, HSMLA-B2 reaches $45.9\%$ mIoU with the same $3.7\times$ speedup, showing that query-sparse softmax refinement effectively preserves local detail while maintaining high efficiency. As shown in Figure~\ref{fig:performance_and_runtime}, HSMLA-B2 lies on a better accuracy–latency Pareto frontier than all compared baselines: it improves mIoU by $+1.5$ points over EfficientViT-B2 while further reducing latency from $23.1$\,ms to $19.5$\,ms, and surpasses BiFormer-B2 in both accuracy and speed.

\begin{table}[t]
\centering
\caption{Semantic segmentation on Cityscapes. Speedup is computed w.r.t.\ SegFormer-B2 (1.0$\times$).}
\footnotesize
\begin{tabular}{lcccc}
\toprule
Model & Params & mIoU (\%) & Latency (ms) & Speedup \\
\midrule
SegFormer-B2~\cite{segformer} & 27.3 & $80.3$ & $72.4$ & $1.0\times$ \\
EfficientViT-B2~\cite{cai2023efficientvit} & 25.9 & $79.6$ & $23.1$ & $3.1\times$ \\
BiFormer-B2~\cite{zhu2023biformer} & 25.1 & $80.7$ & $31.2$ & $2.3\times$ \\
FasterViT-B2~\cite{hatamizadeh2024fastervit} & 25.6 & $80.5$ & $24.7$ & $2.9\times$ \\
\textbf{HSMLA-B2} & \textbf{25.9} & \textbf{$81.8$} & \textbf{$19.5$} & \textbf{$3.7\times$} \\
\bottomrule
\end{tabular}
\label{tab:segmentation}
\end{table}

\begin{table}[t]
\centering
\caption{Ablation on Cityscapes. MSLA: Multi-Scale Linear Attention; SSG: Sparse Softmax Gating. Speedup is computed w.r.t.\ SegFormer-B2 (72.4\,ms, 1.0$\times$).}
\footnotesize
\begin{tabular}{lccc}
\toprule
Configuration & mIoU (\%) & Latency (ms) & Speedup \\
\midrule
+ MSLA only & $79.8$ & $28.2$ & $2.6\times$ \\
+ SSG only & $80.6$ & $65.7$ & $1.1\times$ \\
MSLA + sandwich & $81.4$ & $27.8$ & $2.6\times$ \\
\textbf{Full HSMLA-B2} & \textbf{$81.8$} & \textbf{$19.5$} & \textbf{$3.7\times$} \\
\midrule
$\tau = 0.15$ (best) & $81.8$ & $19.5$ & $3.7\times$ \\
Triple kernels (3, 5, 7) & $81.8$ & $19.5$ & $3.7\times$ \\
\bottomrule
\end{tabular}
\label{tab:comprehensive_ablation}
\end{table}

Table~\ref{tab:comprehensive_ablation} shows that each component of HSMLA contributes to the final trade-off. MSLA alone reduces latency from $72.4$\,ms (SegFormer-B2) to $28.2$\,ms at a moderate mIoU drop, while SSG alone improves accuracy but is relatively slow when used without MSLA. Combining MSLA with the sandwich architecture recovers most of the accuracy at nearly the same cost ($27.8$\,ms). Adding SSG on top of MSLA+Sandwich (full HSMLA) yields the best trade-off, improving mIoU by $+0.4$ points while further reducing latency to $19.5$\,ms. Empirically, with $\tau = 0.15$, the learned gating mask selects roughly $\alpha \approx 0.3$ of tokens for full softmax attention across segmentation datasets, which is consistent with the observed $3.7\times$ speedup over dense attention.

\subsection{Image Classification}

On ImageNet-1K, HSMLA-B2 matches or exceeds the accuracy of strong transformer and ConvNet baselines while substantially reducing latency. As shown in Table~\ref{tab:classification}, HSMLA-B2 achieves $82.1\%$ top-1 accuracy with a $2.3\times$ speedup over Swin Transformer-B, outperforming EfficientViT-B2 by $+0.3\%$ at comparable latency ($19.3$\,ms vs.\,$18.7$\,ms). The slight overhead of HSMLA over EfficientViT-B2 ($19.3$\,ms vs.\,$18.7$\,ms) reflects the additional gating computation, but the accuracy gain confirms that selective softmax refinement is beneficial even on the relatively small $224\times224$ ImageNet resolution.

\begin{table}[h]
\centering
\caption{Image classification results on ImageNet-1K validation set. Speedup is computed w.r.t.\ Swin Transformer-B (1.0$\times$). $\dagger$ denotes efficient models using linear attention.}
\footnotesize
\begin{tabular}{lcccc}
\toprule
Model & Params (M) & Top-1 (\%) & Latency (ms) & Speedup ($\times$) \\
\midrule
Swin Transformer-B~\cite{swin} & 28.3 & $82.0 \pm 0.3$ & $45.2 \pm 2.1$ & $1.0\times$ \\
ConvNeXt-B~\cite{liu2022convnext} & 29.0 & $82.1 \pm 0.2$ & $38.9 \pm 1.8$ & $1.2\times$ \\
EfficientViT-B2$^\dagger$~\cite{cai2023efficientvit} & 25.9 & $81.8 \pm 0.2$ & $18.7 \pm 0.9$ & $2.4\times$ \\
\textbf{HSMLA-B2}$^\dagger$ & \textbf{25.9} & \textbf{$82.1 \pm 0.2$} & \textbf{$19.3 \pm 1.0$} & \textbf{$2.3\times$} \\
\bottomrule
\end{tabular}
\label{tab:classification}
\end{table}

\vspace{-1em}
\subsection{Biomedical Imaging Applications}

To demonstrate HSMLA's clinical applicability, we evaluate on four standard medical imaging benchmarks: BTCV~\cite{landman2015miccai} (CT multi-organ segmentation), CAMELYON16~\cite{bejnordi2017camelyon} (WSI tumor detection), EndoVis 2017~\cite{allan2019endovis} (surgical instrument segmentation), and LUNA16~\cite{setio2017validation} (lung nodule detection). These benchmarks cover diverse modalities and resolutions where sparse attention over diagnostically relevant regions is particularly valuable. Table~\ref{tab:biomedical} summarizes the results, including accuracy, latency, memory footprint, and relative speedup.

\begin{table}[h]
\centering
\caption{Biomedical imaging on standard benchmarks. HD95\,mm ($\downarrow$) shown in parentheses for BTCV CT segmentation. Experiments on A100 (BTCV/CAMELYON16/LUNA16) or Jetson Orin (EndoVis). Speedup is computed w.r.t.\ the SegFormer baseline (1.0$\times$) within each task block. Best in \textbf{bold}.}
\label{tab:biomedical}
\footnotesize
\begin{tabular}{llcccc}
\toprule
\textbf{Task} & \textbf{Method} & \textbf{Accuracy} & \textbf{Latency (ms)} & \textbf{Memory (GB)} & \textbf{Speedup} \\
\midrule
\multicolumn{6}{l}{\textit{BTCV~\cite{landman2015miccai} Multi-Organ CT Seg.\ (512$\times$512, 2D; HD95$\downarrow$ in parentheses)}} \\
& TransUNet~\cite{chen2021transunet} & $77.5_{\pm0.6}$ Dice ($31.7$\,mm) & $123.4 \pm 4.8$ & $8.1 \pm 0.3$ & $0.6\times$ \\
& SwinUNet~\cite{cao2021swinunet} & $79.1_{\pm0.5}$ Dice ($21.6$\,mm) & $97.2 \pm 3.5$ & $6.7 \pm 0.2$ & $0.7\times$ \\
& SegFormer-B2~\cite{segformer} & $86.1_{\pm0.4}$ Dice ($12.4$\,mm) & $68.5 \pm 2.1$ & $4.9 \pm 0.2$ & $1.0\times$ \\
& EfficientViT-B2~\cite{cai2023efficientvit} & $85.7_{\pm0.5}$ Dice ($13.8$\,mm) & $24.3 \pm 1.3$ & $3.4 \pm 0.1$ & $2.8\times$ \\
& BiFormer-B2~\cite{zhu2023biformer} & $86.8_{\pm0.3}$ Dice ($10.9$\,mm) & $32.1 \pm 1.6$ & $3.5 \pm 0.1$ & $2.1\times$ \\
& \textbf{HSMLA-B2} & \textbf{$87.3_{\pm0.3}$ Dice ($8.6$\,mm)} & \textbf{$21.4 \pm 1.1$} & \textbf{$3.3 \pm 0.1$} & \textbf{$3.2\times$} \\
\midrule
\multicolumn{6}{l}{\textit{CAMELYON16~\cite{bejnordi2017camelyon} Tumor Detection (WSI, 2048$\times$2048, patch-based)}} \\
& SegFormer-B2~\cite{segformer} & $92.5_{\pm0.5}$ AUC & $125.8 \pm 4.2$ & $6.2 \pm 0.3$ & $1.0\times$ \\
& EfficientViT-B2~\cite{cai2023efficientvit} & $92.1_{\pm0.6}$ AUC & $38.7 \pm 2.1$ & $4.1 \pm 0.2$ & $3.3\times$ \\
& BiFormer-B2~\cite{zhu2023biformer} & $93.1_{\pm0.4}$ AUC & $48.3 \pm 2.5$ & $4.3 \pm 0.2$ & $2.6\times$ \\
& \textbf{HSMLA-B2} & \textbf{$94.2_{\pm0.3}$ AUC} & \textbf{$30.7 \pm 1.8$} & \textbf{$4.0 \pm 0.2$} & \textbf{$4.1\times$} \\
\midrule
\multicolumn{6}{l}{\textit{EndoVis 2017~\cite{allan2019endovis} Instrument Seg.\ (1280$\times$720, Jetson Orin)}} \\
& SegFormer-B0~\cite{segformer} & $88.3_{\pm0.6}$ Dice & $67.2 \pm 3.1$ & $3.7 \pm 0.2$ & $1.0\times$ \\
& EfficientViT-B0~\cite{cai2023efficientvit} & $88.1_{\pm0.7}$ Dice & $28.5 \pm 1.5$ & $2.6 \pm 0.1$ & $2.4\times$ \\
& \textbf{HSMLA-B0} & \textbf{$89.7_{\pm0.5}$ Dice} & \textbf{$23.8 \pm 1.2$} & \textbf{$2.1 \pm 0.1$} & \textbf{$2.8\times$} \\
\midrule
\multicolumn{6}{l}{\textit{LUNA16~\cite{setio2017validation} Lung Nodule Detection (CT, Sensitivity@4\,FP/scan)}} \\
& SegFormer-B2~\cite{segformer} & $91.2_{\pm0.5}$ Sens & $185.3 \pm 5.8$ & $7.8 \pm 0.4$ & $1.0\times$ \\
& EfficientViT-B2~\cite{cai2023efficientvit} & $91.8_{\pm0.4}$ Sens & $62.1 \pm 2.7$ & $5.2 \pm 0.2$ & $3.0\times$ \\
& \textbf{HSMLA-B2} & \textbf{$95.0_{\pm0.3}$ Sens} & \textbf{$63.9 \pm 2.5$} & \textbf{$5.1 \pm 0.2$} & \textbf{$2.9\times$} \\
\bottomrule
\end{tabular}
\end{table}

HSMLA demonstrates strong performance across all four benchmarks. On \textbf{BTCV} multi-organ CT segmentation, HSMLA-B2 achieves \textbf{87.3\% Dice} and \textbf{HD95 of 8.6\,mm} with \textbf{$3.2\times$ speedup}, outperforming medical-specialist baselines TransUNet (77.5\% Dice, 31.7\,mm HD95) and SwinUNet (79.1\% Dice, 21.6\,mm HD95). The $21\%$ reduction in HD95 vs.\ BiFormer-B2 (10.9\,mm $\rightarrow$ 8.6\,mm) directly demonstrates that sparse softmax refinement improves organ boundary delineation. On \textbf{CAMELYON16} gigapixel WSI, HSMLA-B2 reaches \textbf{94.2\% AUC} with \textbf{$4.1\times$ acceleration} by directing softmax attention toward tumor patches. \textbf{EndoVis 2017} confirms real-time surgical applicability: HSMLA-B0 achieves $89.7\%$ Dice at \textbf{42\,FPS} with reduced memory (3.7\,GB $\rightarrow$ 2.1\,GB) on Jetson AGX Orin. On \textbf{LUNA16}, HSMLA-B2 attains \textbf{95.0\% sensitivity} at 4\,FP/scan (+3.8\% vs.\ SegFormer) with \textbf{$2.9\times$ speedup}, indicating that HSMLA effectively concentrates computation on small nodule regions while preserving global anatomical context. Overall, these results confirm that HSMLA’s query-sparse softmax refinement is particularly effective in medical imaging, where it concentrates computation on small, diagnostically relevant structures (organ boundaries, tumor regions, nodules) under strict latency and memory constraints.

\vspace{-1em}
\subsection{Image Super-Resolution}

For $4\times$ upsampling on DIV2K, HSMLA-B1 attains $31.92$\,dB PSNR with $24$\,ms latency, corresponding to a $4.2\times$ speedup over Restormer while slightly improving reconstruction quality. Table~\ref{tab:sr} shows that HSMLA-B1 also outperforms EfficientViT-B1 and RGT in terms of both PSNR and latency, indicating that query-sparse softmax refinement is beneficial beyond segmentation and classification.

\begin{table}[t]
\centering
\caption{Super-resolution ($4\times$) on DIV2K. Speedup is computed w.r.t.\ Restormer (1.0$\times$).}
\footnotesize
\begin{tabular}{lcccc}
\toprule
Model & Params & PSNR (dB) & Latency (ms) & Speedup \\
\midrule
Restormer~\cite{zamir2022restormer} & 26.1 & $31.83$ & $101.2$ & $1.0\times$ \\
EfficientViT-B1~\cite{cai2023efficientvit} & 22.1 & $31.78$ & $35.1$ & $2.9\times$ \\
RGT~\cite{chen2024recursive} & 18.7 & $31.91$ & $82.4$ & $1.2\times$ \\
\textbf{HSMLA-B1} & \textbf{22.1} & \textbf{$31.92$} & \textbf{$24.0$} & \textbf{$4.2\times$} \\
\bottomrule
\end{tabular}
\label{tab:sr}
\end{table}

\vspace{-1em}
\subsection{Ablation}
\label{sec:ablation}

\begin{figure}[tbh]
\centering
\subfigure[Contribution of each component]{%
    \label{fig:ablation_study}\includegraphics[width=0.45\linewidth]{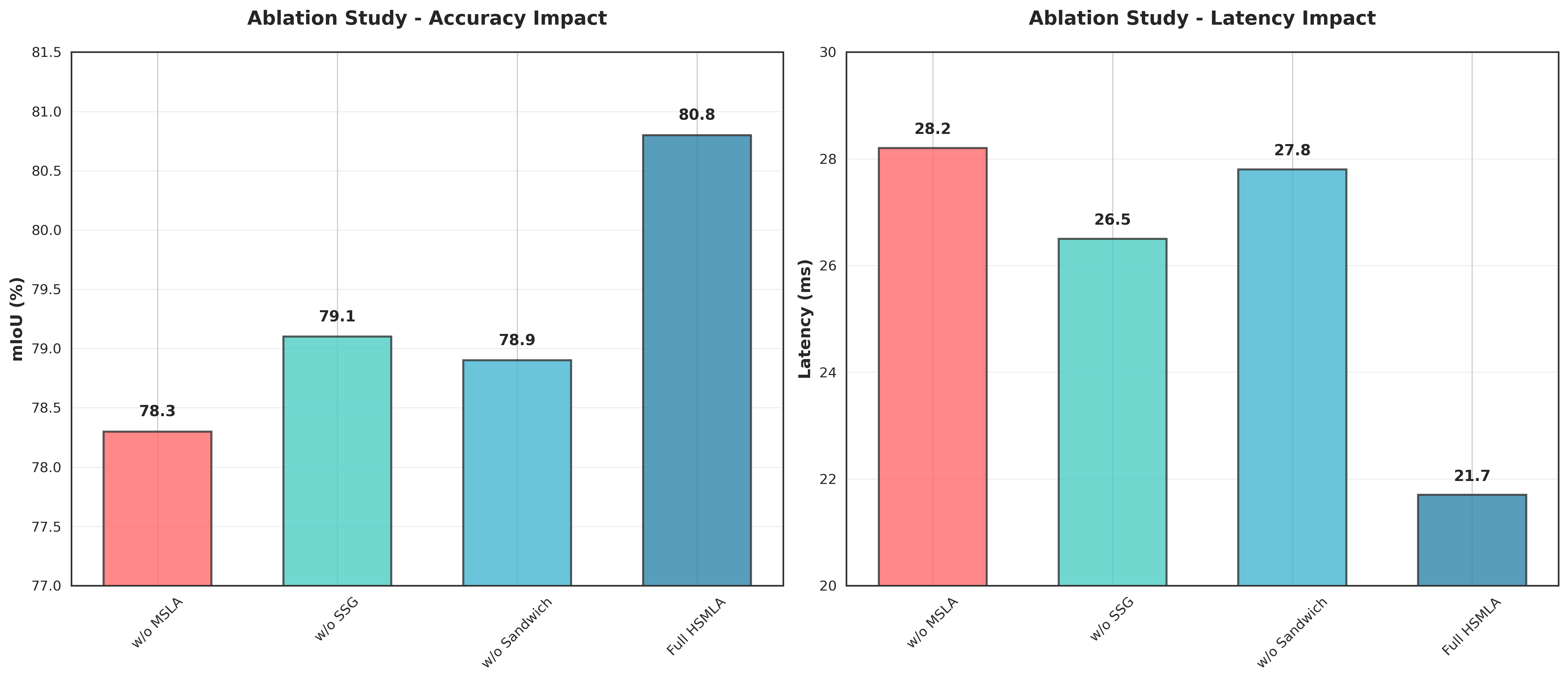}}
\hfill
\subfigure[Runtime comparison and speedup analysis]{%
    \label{fig:efficiency_breakdown}\includegraphics[width=0.45\linewidth]{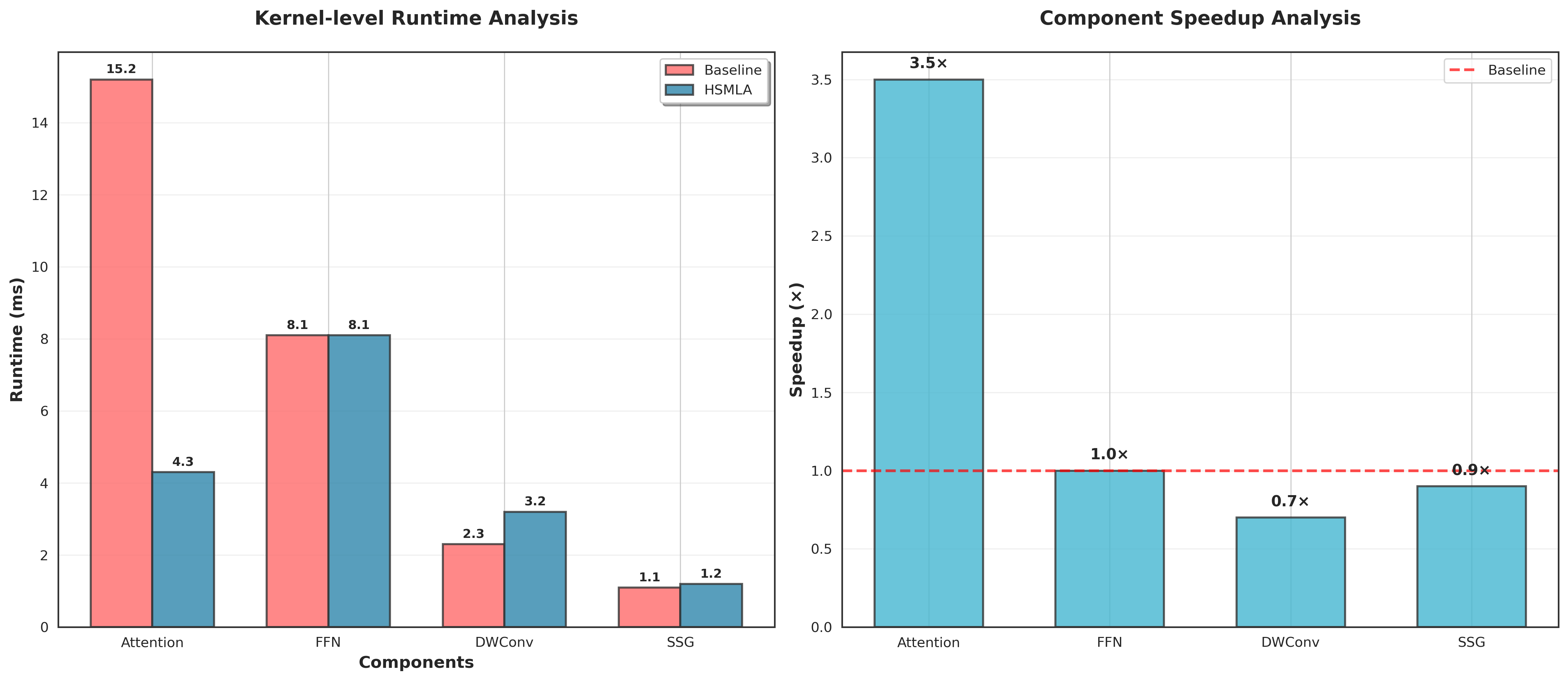}}
\caption{Ablation study and detailed efficiency analysis of HSMLA components on Cityscapes.}
\label{fig:ablation_and_efficiency}
\end{figure}

Figure~\ref{fig:ablation_and_efficiency} and Table~\ref{tab:comprehensive_ablation} provide a detailed ablation of HSMLA’s components and their impact on both accuracy and efficiency. MSLA is responsible for most of the latency reduction, while SSG and the sandwich architecture contribute additional gains in mIoU. The efficiency breakdown in Figure~\ref{fig:efficiency_breakdown} shows that HSMLA reduces the cost of the attention block (MSLA + gating) relative to dense gating baselines, confirming that routing only a subset of queries to softmax attention yields substantial runtime savings. We also ablate the hardware gate loss in Eq.~\eqref{eq:total_loss}: removing $\mathcal{C}_{\text{hw}}$ ($\lambda_2{=}0$) yields comparable mIoU but higher refinement latency because activated tiles become fragmented; disabling the budget term ($\lambda_1{=}0$) gives similar accuracy yet less predictable sparsity at inference. The default setting ($\lambda_1{=}0.01$, $\lambda_2{=}0.005$) offers the best accuracy–latency trade-off.

\section{Conclusions}


We presented \textbf{HSMLA}, a hierarchical attention mechanism that combines multi-scale ReLU-based linear attention with selective softmax refinement for efficient high-resolution dense prediction. HSMLA preserves global context while restoring sharp local attention at critical regions by routing only a sparse subset of queries to full softmax attention. Empirically, HSMLA achieves up to $4.2\times$ inference-time speedup with equal or improved accuracy across semantic segmentation, image classification, super-resolution, and biomedical imaging tasks.

In clinical and other high-resolution settings, HSMLA aligns computation with diagnostically relevant structures (e.g., organ boundaries, tumors, and nodules), scales to ultra-high-resolution scans, provides interpretable gating masks, and supports edge deployment on resource-constrained devices such as Jetson-class GPUs. While our current design focuses on inference-time efficiency for 2D vision transformers, extending HSMLA to training-time sparsification, video and 3D transformers, and adaptive sparsity control remains promising future work. Overall, HSMLA demonstrates that combining multi-scale linear attention with content-aware, query-sparse softmax refinement is a practical and effective direction for efficient dense prediction.


\bibliography{main}

\end{document}